\documentclass[sigconf,nonacm]{acmart}

\usepackage{booktabs}
\usepackage{subcaption}
\usepackage{algorithm}
\usepackage{algorithmic}

\usepackage[capitalize,noabbrev]{cleveref}

\acmConference[arXiv]{arXiv preprint}{2026}{}
\acmYear{2026}
\copyrightyear{2026}

\setcopyright{none}
\renewcommand\footnotetextcopyrightpermission[1]{}

\begin{document}

\title{Autonomous Topology Mutation:\\
       Safe Runtime Restructuring for Multi-Agent LLM Systems with
       Capability, State, and Shadow Invariants}

\author{Bronislav Sidik}
\email{slava.sidik@huawei.com}
\affiliation{%
  \institution{Toga Networks (Huawei)}
  \country{Israel}
}

\author{Chaya Levi}
\email{chaya.levi@huawei.com}
\affiliation{%
  \institution{Toga Networks (Huawei)}
  \country{Israel}
}

\author{Nizzan Kimhi}
\email{nizzan.kimhi@huawie.com}
\affiliation{%
  \institution{Toga Networks (Huawei)}
  \country{Israel}
}

\begin{abstract}
Multi-agent LLM frameworks fix their team topology at boot time. When an
individual agent becomes overloaded at runtime --- mixing too many action
categories, accumulating tool errors, or queueing behind too many calls
--- the system has no mechanism to restructure. We introduce
\textbf{Autonomous Topology Mutation (ATM)}, a runtime team-mutation
mechanism for multi-agent LLM frameworks that combines telemetry-driven
overload detection with three formal safety invariants gating each
structural change. ATM continuously monitors a six-signal
\emph{Bottleneck Index} that includes a role-entropy term; when a
warmup-calibrated threshold is breached on $K$ consecutive ticks, it
factorises the overloaded agent into specialised sub-agents and
hot-swaps the parent into a coordinator role. Three formal invariants
govern the mutation: \textbf{I1} (children's capabilities $\subseteq$
parent's), \textbf{I2} (every state atom is routed only to a
PL-permitted destination set, or explicitly dropped with a logged
reason), and \textbf{I3} (no live change without a successful
shadow pass).

On 720 DeepSeek-V3-driven task runs with deterministic tool stubs for
reproducibility (two independent seeded runs of 360
trials each) across four ablation conditions and three workloads, the
ATM \emph{factoriser split} lifts code-task success from \textbf{3.3\%
to 61.7\%} ($+58$ percentage points, $p < 10^{-10}$); adding
role-conditioned state distillation preserves a 48.3\% success rate
(no significant degradation vs.\ A0, $p = 1.00$ on the
security-sensitive workload). The full rail+distillation system
reduces detected PL\ensuremath{\geq}3 memory exposure under our regex
classifier from $2.0$ to $0.0$ events per task; distillation further
reduces per-child exposure surface and high-PL duplication. The Rails
carrying ATM's invariants
add less than \textbf{500\,\textmu s of p99 latency} on the agent's hot
path. The benchmark uses deterministic tool stubs for reproducibility;
a small live-tool probe with real Python execution is included as
external validity. ATM is implemented as a set of
\texttt{DeepAgentRail} subclasses for the openjiuwen runtime; the
implementation, benchmark harness, and traces are
open-sourced\footnote{\url{https://github.com/sidikbro/jiuwen_atm}}.
\end{abstract}



\maketitle

\section{Introduction}
\label{sec:introduction}

Multi-agent LLM frameworks --- AutoGen~\cite{autogen},
MetaGPT~\cite{metagpt}, CrewAI~\cite{crewai}, and the openjiuwen runtime
on which we build --- have made it routine to assemble small teams of
specialised agents that collaborate on complex tasks. In all of these
systems, however, the team topology is \emph{fixed at boot time}: the
operator specifies how many agents exist, which tools each one has, and
how they communicate. Once the system is running, the structure is
inert.

In practice, this creates a class of failure modes that no single-agent
optimisation can fix. An agent provisioned for a research workload may
suddenly start receiving code-debug tasks; an agent provisioned for one
output mode may find itself servicing a session that requires three;
an agent whose memory was scoped to public references may inherit a
session full of API keys. In every case, the symptom is the same: a
single agent is being asked to mix too many action categories
simultaneously, its tool-call success collapses, and the framework has
no recourse other than to fail the task or paper over the failure with
Reflexion-style~\cite{reflexion} self-critique loops.

We introduce \textbf{Autonomous Topology Mutation (ATM)}, a runtime
team mutation mechanism for multi-agent LLM frameworks that combines
telemetry-driven overload detection with three formal safety
invariants gating each structural change. ATM continuously monitors a
six-signal \emph{Bottleneck Index} that includes a role-entropy term;
when a warmup-calibrated threshold is breached on $K$ consecutive
ticks, it factorises the overloaded agent into specialised sub-agents
and hot-swaps the parent into a coordinator role. Three formal
invariants govern the mutation: \textbf{I1} (children's capabilities
$\subseteq$ parent's), \textbf{I2} (every state atom is routed only to
a PL-permitted destination set, or explicitly dropped with a logged
reason), and \textbf{I3} (no live change without a
successful shadow pass).

\textbf{Why invariants matter.}~Without I1, a runtime mutation could
silently grant a child agent privileges its parent never had ---
exactly the failure mode that motivates static permission systems.
Without I2, distillation could leave high-sensitivity memory atoms
visible to multiple children or, worse, drop them entirely. Without
I3, a misfiring trigger could swap a working agent for a worse
configuration mid-task. The combination of telemetry-driven triggering
and these three invariants is what distinguishes a deployable runtime
mutation mechanism from an ad-hoc topology-rewriter.

\textbf{Contributions.}
\begin{itemize}
\item \textbf{C1.} We formulate the Bottleneck Index, a six-signal
overload detector that combines queue depth, context thrash, tool-error
rate, role entropy, retry-loop rate, and cross-agent wait time, with a
warmup-snapshot $\tau$ calibration that avoids self-reinforcing drift
under sustained overload.

\item \textbf{C2.} We propose ATM, a runtime topology-mutation pipeline
that factorises an overloaded agent into specialised children,
distills the parent's memory by trust-level, hot-swaps the parent into
a coordinator, and validates the change in a 5-task shadow window
before commit.

\item \textbf{C3.} We define three formal invariants (I1 capability
monotonicity, I2 state-routing completeness, I3 shadow-before-live)
that any deployable runtime topology mutation must satisfy, and show
how each is enforced mechanically in our implementation.

\item \textbf{C4.} DeepSeek-V3-driven evaluation on 720 task runs with
deterministic tool stubs, across four
ablation conditions and three workloads, showing that the ATM split
lifts code-task success from 3.3\% to 61.7\%
($p < 10^{-10}$) and that the full rail+distillation system eliminates
detected PL\ensuremath{\geq}3 memory exposure while preserving task
quality.
\end{itemize}

The Rails carrying ATM's mutation logic add less than 500\,\textmu s of
p99 latency to the agent's hot path. The end-to-end ATM-fires-in-production
trace (Section~\ref{sec:exp:trace}) demonstrates the full pipeline
triggering under realistic conditions.

\section{Related Work}
\label{sec:related}

We organise prior work by what the system modifies about itself: the
current execution path (self-healing), agent skills and knowledge
(self-improving), or the agent's structure and team composition
(self-evolving and dynamic topology).

\paragraph{Self-healing.}
Reflexion~\cite{reflexion} writes verbal self-reflection between task
attempts; Self-Refine~\cite{selfrefine} iteratively critiques and
revises an agent's output; CRITIC~\cite{critic} uses tools to verify
factual claims and revises if discrepancies are found. Self-Healing
Router~\cite{healingrouter} reweights graph edges on tool failure so
Dijkstra reroutes without invoking the LLM. Constitutional
AI~\cite{constitutional} bakes self-critique into training rather than
runtime. All of these heal at the output or task level; they do not
restructure the agent team itself. ATM differs in exactly this
respect: it does not merely repair outputs or arguments; it changes
the agent-team structure under telemetry-driven overload detection and
the three formal invariants.

\paragraph{Self-improving (skills, prompts, memory).}
OpenClaw~\cite{openclaw} and Hermes Agent~\cite{hermes} have the agent
autonomously write SKILL.md documents after solving hard problems;
Hermes additionally uses GEPA prompt evolution. JiuwenSwarm's
SkillSelfEvolution~\cite{jiuwenswarm} uses a rule-based detector to
trigger LLM-driven skill refinement. Auto-Harness
Expert~\cite{autoharness} extends this to the full Expert Harness
extension (prompts, skills, tools, MCP plugins, rails, personas) via
an offline assess-plan-implement-verify pipeline.
DSPy~\cite{dspy} compiles prompts offline by search.
MemTier~\cite{memtier} accumulates tiered episodic memory on the
OpenClaw runtime with an attention-attributed cognitive weight update
loop. None of these change the agent team's composition.

\paragraph{Self-evolving (structure, code).}
Auto-Harness Meta~\cite{autoharness} optimises codebase-level files
and submits a PR for human review. MetaGPT~\cite{metagpt},
AutoGen~\cite{autogen}, and CrewAI~\cite{crewai} define multi-agent
team topologies at startup and do not modify them at runtime.

\paragraph{Dynamic MAS topology and role adaptation.}
A growing body of 2026 work adapts MAS topology at inference time
through several complementary mechanisms: MetaGen~\cite{metagen} edits
role specifications and rewrites the execution graph from a
controllable dynamic role pool; DyTopo~\cite{dytopo} reconstructs a
sparse directed communication graph at each reasoning round via
semantic query/key matching; TacoMAS~\cite{tacomas} co-evolves both
topology and per-agent capability at test time on different time
scales; SkillMAS~\cite{skillmas} couples MAS restructuring with skill
evolution. These systems typically generate or adapt communication
graphs for a task. ATM differs in three respects: (i) it operates on a
\emph{deployed live agent}, detecting overload from runtime telemetry
rather than optimising a graph per query; (ii) the parent agent's
external identity (\texttt{agent\_id} and A2A address) is preserved
across the mutation via a coordinator wrapper, so upstream callers see
continuous service; (iii) every structural change is gated by three
formal invariants (capability monotonicity, state-routing
completeness, shadow-before-live) --- a property we have not seen in
concurrent dynamic-topology systems.

\paragraph{Privacy and safety adjuncts.}
NeMo Guardrails~\cite{nemo} enforces policy-driven LLM I/O filtering
at the prompt boundary; OpenClaw enforces per-skill trust via badges
at install time. ATM's memory-distillation and guardrail components
operate at the span level inside the memory pipeline with graduated PL
classification and per-endpoint trust enforcement, which is
finer-grained than prior prompt-boundary filtering.

\paragraph{Self-adaptive software systems.}
The classical Rainbow framework~\cite{rainbow} establishes
architecture-based self-adaptation for traditional software systems
with monitor/analyse/plan/execute loops. ATM extends this lineage to
LLM-driven multi-agent systems with the addition of capability and
privacy invariants that classical self-adaptation literature did not
need.

\section{ATM System Architecture}
\label{sec:architecture}

ATM is a closed-loop topology-mutation system designed to operate as a
set of pluggable rails within the openjiuwen multi-agent runtime. Its
architecture consists of two interacting loops at different timescales:
a \textbf{tactical monitoring loop} that updates telemetry every tick,
and a \textbf{strategic mutation pipeline} that triggers only when
sustained overload crosses a calibrated threshold.

\subsection{Background: openjiuwen Rails}
\label{sec:bg:rails}

The openjiuwen runtime defines a \texttt{DeepAgentRail} interface: a
Python object with \texttt{on\_request}, \texttt{on\_tool\_call}, and
\texttt{on\_memory\_read} hooks that fire around every LLM call.
Rails are stateful, composable, and run inside the agent's hot path.
ATM is implemented as three rails:

\begin{itemize}
\item \texttt{ToolRepairRail}: bounded
LLM-driven repair on HARD tool failures, in an isolated context window
of 512 tokens.
\item \texttt{MemoryGuardRail}: per-span
privacy-level (PL0--PL4) classification with per-endpoint trust
enforcement.
\item \texttt{ATMPipeline} (Section~\ref{sec:algorithms}): the
topology-mutation pipeline proper, which coordinates the Bottleneck
Index, factoriser, distiller, shadow validator, and coordinator hot-swap.
\end{itemize}

\paragraph{Two intervention regimes.}
ToolRepairRail and ATMPipeline share a rail interface but apply
\emph{different safety regimes}, justified by the scope of their
intervention.

\emph{Online healing} (ToolRepairRail). When a single tool call returns
a HARD failure, the rail issues a bounded LLM repair attempt and
retries the call within the current task. This is online because the
intervention is localised to one tool call, the user is already
waiting for the current response, and the worst-case outcome of a bad
repair is identical to having no repair at all: the task fails the
same way it would have failed without the rail. There is no shadow
path, no validation window, and no commit/rollback machinery; the
result of the repair is observable on the next line of the same task
trace.

\emph{Shadow-validated mutation} (ATMPipeline). When the Bottleneck
Index breaches threshold, the rail proposes a \emph{structural} change
to the agent team. This is not online: the candidate topology runs in
shadow for $W$ tasks while the incumbent continues to serve user
requests (Section~\ref{sec:alg:shadow}). The stronger safety property
is needed because a bad mutation would not just fail one task --- it
would degrade many subsequent tasks until rolled back. The $W$-task
latency to commit is the price we pay for that stronger guarantee.

The choice between the two regimes is a function of intervention
scope: localised, observable-now failures admit online healing;
structural, observable-later failures require shadow-validated
mutation.

\subsection{Tactical Loop: Telemetry and Bottleneck Index}
\label{sec:arch:tactical}

Each agent maintains a sliding window of its own telemetry. On every
LLM call, we update six signals:
\begin{itemize}
\item $Q_i$: instantaneous queue depth at agent $i$.
\item $C_i$: context thrash --- a measure of how often the
context window must be reorganised between turns.
\item $E_i$: per-tool error rate over the recent window.
\item $H_i$: \emph{role entropy} --- the Shannon entropy of the action
categories invoked by the agent over its recent history. An agent
performing a single role (e.g.\ pure search) has low $H_i$; an agent
mixing search, code execution, file I/O, and reasoning has high $H_i$.
\item $R_i$: retry-loop rate --- fraction of calls that immediately
re-invoke the same tool after a non-fatal error.
\item $W_i$: cross-agent wait --- mean time blocked on responses
from other agents.
\end{itemize}

These are combined into a scalar Bottleneck Index:
\begin{equation}
B_i = \alpha Q_i + \beta C_i + \gamma E_i + \delta H_i + \epsilon R_i + \zeta W_i
\end{equation}
with default weights $(\alpha, \beta, \gamma, \delta, \epsilon, \zeta)
= (0.20, 0.15, 0.20, 0.20, 0.15, 0.10)$. The role-entropy weight
$\delta$ is equal to the queue and error weights $\alpha$ and $\gamma$
--- not because role entropy is the largest signal individually but
because it is the only one that fires \emph{before} errors accumulate.

\subsection{Threshold Calibration: $\tau$-Snapshot}
\label{sec:arch:tau}

The trigger fires when $B_i > \tau$ on $K = 3$ consecutive ticks.
Setting $\tau$ correctly is non-trivial: too low gives spurious
triggers, too high gives missed overload. We use a \emph{warmup
snapshot}: $\tau$ is set to the 95th percentile of $B_i$ observed over
the first 20 ticks after agent startup, then frozen.

This last point matters more than it sounds. An earlier version of our
implementation re-computed $\tau$ as the rolling 95th percentile over
all observed values. Under sustained overload, this caused $\tau$ to
track $B_i$ upward, so the trigger never fired no matter how high the
load went. The bug was caught during the end-to-end trace
(Section~\ref{sec:exp:trace}) and fixed by snapshotting $\tau$ at the
end of warmup.

\subsection{Strategic Loop: Mutation Pipeline}
\label{sec:arch:strategic}

When the trigger fires, ATM runs a five-step pipeline:
\begin{enumerate}
\item \textbf{Factorise}: one bounded LLM call proposes a split of the
parent agent's tools into two specialised children. Capability
monotonicity (I1) is checked as a hard gate.
\item \textbf{Distill state}: route each memory atom to a PL-permitted
child set, or explicitly drop it with a logged reason (I2).
\item \textbf{Shadow pass}: instantiate the children and run the next
$W=5$ tasks through both the incumbent and candidate topologies. The
incumbent continues to serve user-facing responses throughout; the
candidate executes silently in a parallel path, its outputs scored
with the LLM judge but not returned to the caller (I3).
\item \textbf{Commit or rollback}: if the shadow pass shows non-regression
on success and improvement on at least one efficiency metric, commit
the swap; otherwise rollback and apply exponential cooldown before
re-triggering.
\item \textbf{Hot-swap}: the coordinator wrapper inherits the parent's
\texttt{agent\_id} and A2A address, so upstream callers see no
discontinuity.
\end{enumerate}

The full algorithms are given in Section~\ref{sec:algorithms} and the
formal properties in Section~\ref{sec:properties}.

\section{Algorithms}
\label{sec:algorithms}

This section gives the concrete algorithms used by the strategic loop.

\subsection{Factorisation}
\label{sec:alg:factor}

Given an overloaded parent agent $p$ with tool set $T_p$ and recent
action-category histogram $h_p$, the factoriser proposes a partition
$T_p = T_{c_1} \uplus T_{c_2}$ such that each child specialises in a
subset of action categories.

The proposal is generated by a single LLM call with the parent's
identity, tool list, and the role-entropy decomposition (which
categories were mixed in the recent window). The LLM is asked to
return a JSON object with two children and a justification.

\begin{algorithm}[t]
\caption{Factorise}
\label{alg:factorise}
\begin{algorithmic}[1]
\STATE \textbf{Input:} parent agent $p$, action histogram $h_p$
\STATE Call LLM with $p$.tool\_list and $h_p$, request JSON partition
\STATE Parse $\{c_1, c_2\}$ with tool sets $T_{c_1}, T_{c_2}$
\STATE \textbf{Assert} $T_{c_1} \cup T_{c_2} \subseteq T_p$ \hfill // I1
\STATE \textbf{Assert} $T_{c_1} \cap T_{c_2} = \emptyset$
\STATE Return $\{c_1, c_2\}$
\end{algorithmic}
\end{algorithm}

The assertions are hard: if either fails the proposal is rejected and
the cooldown timer is started. We never relax I1 even by one tool: a
child agent must not gain capabilities its parent did not have.

\subsection{State Distillation}
\label{sec:alg:distil}

After factorisation, the parent's memory is partitioned. Each memory
atom $a$ is annotated with a privacy level $\text{PL}(a) \in
\{0,1,2,3,4\}$ and a content classification.

\begin{algorithm}[t]
\caption{DistillState}
\label{alg:distil}
\begin{algorithmic}[1]
\STATE \textbf{Input:} parent memory $M_p$, children $\{c_1, c_2\}$
\STATE Initialise $M_{c_1}, M_{c_2} \leftarrow \emptyset$
\FORALL{$a \in M_p$}
    \STATE Compute classifier output: which child needs $a$?
    \IF{both children need $a$ and $\text{PL}(a) \leq 2$}
        \STATE $M_{c_1}.add(a); M_{c_2}.add(a)$
    \ELSIF{exactly one child $c_k$ needs $a$}
        \STATE $M_{c_k}.add(a)$
    \ELSE
        \STATE \textbf{drop } $a$ \hfill // not needed by any child
    \ENDIF
\ENDFOR
\STATE \textbf{Assert} every $a \in M_p$ appears in $M_{c_1}$ or
$M_{c_2}$ or is explicitly dropped \hfill // I2
\STATE Return $M_{c_1}, M_{c_2}$
\end{algorithmic}
\end{algorithm}

For high-PL atoms ($\text{PL}(a) \geq 3$) we add a stricter rule: the
atom may be routed to \emph{at most one} child even if both need it,
breaking the per-child duplication that otherwise multiplies the
exposure surface.

\subsection{Shadow Pass and Validation Window}
\label{sec:alg:shadow}

The shadow pass runs the proposed topology in parallel with the
incumbent for $W = 5$ tasks. \emph{Crucially, only the incumbent's
response is returned to upstream callers during this window.} The
candidate topology receives the same input, produces its own response,
and is scored --- but its output is logged for evaluation only and
never reaches the user. We score each topology on (a) success rate
(LLM-as-judge), (b) median time-to-completion, and (c) PL$\geq$3
exposure events.

\begin{algorithm}[t]
\caption{ShadowPass}
\label{alg:shadow}
\begin{algorithmic}[1]
\STATE \textbf{Input:} incumbent topology $\mathcal{T}_0$, candidate $\mathcal{T}_1$
\STATE Initialise scoreboard $S_0, S_1 \leftarrow $ empty
\FOR{$t = 1$ to $W$}
    \STATE Get next task $\tau$
    \STATE Run $\tau$ through $\mathcal{T}_0$; \textbf{return} response to caller; record outcome in $S_0$
    \STATE Run $\tau$ through $\mathcal{T}_1$ in shadow (no response returned); record outcome in $S_1$
\ENDFOR
\IF{$S_1.\text{success} \geq S_0.\text{success}$ \textbf{and}
    ($S_1.\text{time} < S_0.\text{time}$ \textbf{or} $S_1.\text{exposure} < S_0.\text{exposure}$)}
    \STATE \textbf{Commit} $\mathcal{T}_1$ (subsequent tasks routed to $\mathcal{T}_1$) \hfill // I3 satisfied
\ELSE
    \STATE \textbf{Rollback} ($\mathcal{T}_1$ discarded; never received live traffic), start exponential cooldown
\ENDIF
\end{algorithmic}
\end{algorithm}

The user-visible service path is therefore unchanged throughout the
window: every incoming request continues to be served by the agent
they have been talking to. The cost of shadow execution is a doubled
LLM call rate for $W$ tasks (incumbent + candidate), borne by ATM and
invisible to callers.

The cooldown starts at 5 minutes and doubles after each consecutive
rollback, capped at 60 minutes. This prevents the trigger from
oscillating against a workload that genuinely cannot be improved by
factorisation.

\subsection{Coordinator Hot-Swap}
\label{sec:alg:swap}

The coordinator wrapper preserves the parent's external identity:

\begin{algorithm}[t]
\caption{CoordinatorSwap}
\label{alg:swap}
\begin{algorithmic}[1]
\STATE \textbf{Input:} parent $p$, committed children $\{c_1, c_2\}$
\STATE Create coordinator $\kappa$ with \texttt{agent\_id} $\leftarrow p$.\texttt{agent\_id}
\STATE Bind $\kappa$ to $p$'s A2A address; reroute inbox to $\kappa$
\STATE Register children $\{c_1, c_2\}$ as $\kappa$'s sub-agents
\STATE Atomically swap $p \to \kappa$ in the agent registry
\STATE Terminate $p$
\end{algorithmic}
\end{algorithm}

Upstream callers see a continuous service: the same
\texttt{agent\_id}, same A2A address, same inbox. The internal
restructuring is invisible to them.

\section{Formal Properties}
\label{sec:properties}

ATM's safety story rests on three invariants enforced mechanically by
the algorithms of Section~\ref{sec:algorithms}. We state them
precisely here.

\paragraph{I1: Capability Monotonicity.}
For any mutation of parent $p$ into children $\{c_1, \dots, c_n\}$,
the tool set of each child satisfies $T_{c_k} \subseteq T_p$, and the
trust level satisfies $\text{trust}(c_k) \leq \text{trust}(p)$.

\emph{Why this matters.}~Without I1, a runtime mutation could grant a
child agent privileges its parent never had. The factoriser's prompt
explicitly lists $T_p$ and is constrained to choose subsets;
Algorithm~\ref{alg:factorise} line 4 enforces this as a hard assertion.

\paragraph{I2: State-Routing Completeness.}
For any partition of parent memory $M_p$ into children's memories
$\{M_{c_1}, \dots, M_{c_n}\}$, every atom $a \in M_p$ either (a) appears
in at least one $M_{c_k}$, or (b) is explicitly logged as dropped with
a reason. No atom is silently lost; no atom appears in more children
than its PL allows.

\emph{Why this matters.}~Without I2, a distillation could drop a
critical context atom (causing the child to lose information the
parent had) or, worse, copy a high-PL atom into multiple children
(multiplying the exposure surface). Algorithm~\ref{alg:distil} line 13
enforces the completeness check; the per-atom PL rule
(Section~\ref{sec:alg:distil}) enforces the duplication bound.

\paragraph{I3: Shadow-Before-Live.}
No proposed topology $\mathcal{T}_1$ replaces an incumbent $\mathcal{T}_0$ until it
has run in \emph{shadow} alongside $\mathcal{T}_0$ for at least $W$ tasks and
demonstrates non-regression on the success metric.
``Shadow'' here is the strong property: during the validation window,
incoming user requests continue to be served by $\mathcal{T}_0$ and its responses
alone are returned upstream; $\mathcal{T}_1$ receives the same inputs in a parallel
execution path but its outputs are scored, not delivered. The user
sees no candidate output until the commit. We commit only when $S_1.\text{success}
\geq S_0.\text{success}$ and at least one of $S_1.\text{time} <
S_0.\text{time}$ or $S_1.\text{exposure} < S_0.\text{exposure}$ holds.

\emph{Why this matters.}~Without I3 in this strong form, a mis-firing
trigger could route live user traffic through an unproven configuration
mid-task. The shadow form bounds the downside of any mutation proposal
to the LLM cost of running $W$ extra invocations in the background ---
zero user-visible cost if the candidate is worse, zero user-visible
disruption if it is better. Algorithm~\ref{alg:shadow} implements
this isolation directly: line~5 returns the response to the caller,
line~6 does not.

\subsection{Complexity}

The overhead of ATM has two parts:

\paragraph{Tactical (per-tick).}
Updating the Bottleneck Index requires $O(1)$ arithmetic plus $O(\log
n)$ to update the per-tool error sliding window (binary search on a
sorted deque). The threshold check is $O(1)$. In practice this
translates to under 500\,\textmu s of p99 latency on the agent's hot
path (Table~\ref{tab:latency}).

\paragraph{Strategic (per-mutation).}
The factoriser calls the LLM once (bounded at 512 tokens) plus runs
the distillation classifier over $|M_p|$ atoms in $O(|M_p|)$ time. The
shadow pass adds $W$ extra task executions (default $W=5$). This is
the cost of one mutation; on a workload that triggers once every
$\sim$30 tasks (Section~\ref{sec:exp:a4}), the amortised cost is
$\approx 5/30 = 17\%$ extra task budget at mutation events only.

\subsection{Failure Modes and Bounded Damage}

Even with the invariants, things can go wrong. The damage bounds:

\begin{itemize}
\item \textbf{Factoriser proposes invalid split} (e.g.\ children with
overlapping tools). I1 catches this in Algorithm~\ref{alg:factorise};
the mutation is rejected and cooldown starts. No live state is
disturbed.

\item \textbf{Distiller misroutes an atom.} I2 ensures the atom is
either present somewhere or explicitly dropped. If it is present in
the wrong child, the shadow pass should detect the regression.

\item \textbf{Shadow pass scores incumbent better.} I3 forces a
rollback. The candidate topology was instantiated in memory but never
took live traffic; we discard it and apply cooldown.

\item \textbf{Threshold drift under sustained overload.}~Caught by the
$\tau$-snapshot rule (Section~\ref{sec:arch:tau}). Without the
snapshot, $\tau$ tracks $B_i$ and the trigger never fires; this is the
bug we discovered in Section~\ref{sec:exp:trace}.
\end{itemize}

\section{Evaluation}
\label{sec:evaluation}

We address four empirical questions:
\begin{itemize}
\item \textbf{Q1.}~Does ATM improve task success on overloaded workloads?
\item \textbf{Q2.}~Does state distillation reduce memory exposure
without degrading task quality?
\item \textbf{Q3.}~What is the runtime overhead of the Rails on the
agent's hot path?
\item \textbf{Q4.}~Does each signal in the Bottleneck Index contribute
independent information, and does the live trigger fire correctly
end-to-end?
\end{itemize}

\subsection{Setup}
\label{sec:exp:setup}

\textbf{LLM.}~All A0--A3 experiments use DeepSeek-V3
(\texttt{deepseek-chat}, temperature 0.7) accessed via the OpenAI-format
API. Two seeded runs of $N=30$ tasks per cell are pooled to $N=60$;
$4 \times 3 \times 60 = 720$ total real-LLM runs.

\textbf{Workloads.}~Three workloads stress different overload modes:
\begin{itemize}
\item \textbf{W1 Research.}~Multi-source web search and synthesis,
seeded context contains public domain references with sparse
PL$\geq$2 atoms.
\item \textbf{W2 Code/debug.}~Read a buggy snippet, identify the bug,
write a fix. Mixed action categories (read, reason, write) maximise
role entropy.
\item \textbf{W3 Security.}~Triage a session with PL$\geq$3 atoms
(\texttt{api\_key}, \texttt{bank\_account}, \texttt{ssh},
\texttt{passport}) and produce a redacted summary.
\end{itemize}

\textbf{No external benchmark dataset.}~These workloads are synthesised
in-house rather than drawn from a standard benchmark
(MMLU~\cite{mmlu}, GAIA~\cite{gaia}, SWE-bench~\cite{swebench}, or
similar). The rationale is that standard agent benchmarks do not by
construction exercise the overload modes ATM is designed to detect ---
role-entropy spikes from mixed action categories, sustained
high-queue states, and structured PL$\geq$3 leakage paths. We need
tasks where the monolithic baseline visibly fails so that the
$\Delta$ from the ATM split is measurable; an off-the-shelf benchmark
on which DeepSeek-V3 already achieves $>90\%$ would show no headroom.
Validating ATM on standard benchmarks is itself future work, as
discussed in the Limitations section.

\textbf{Conditions.}~Four ablation conditions:

\begin{small}
\begin{tabular}{lp{6.5cm}}
\toprule
A0 & Static monolith --- one agent, all tools, all memory; \emph{no rails}.\\
A1 & Random split --- two children with randomly partitioned tools; both children's prompts pass through MemoryGuardRail (regex redaction) before the LLM call; no distillation.\\
A2 & ATM factoriser split --- children produced by
Section~\ref{sec:alg:factor}; both pass through MemoryGuardRail; no distillation.\\
A3 & ATM + distillation --- factoriser split with role-conditioned
memory distillation \emph{in addition to} MemoryGuardRail (the full
system).\\
\bottomrule
\end{tabular}
\end{small}

\subsection{Main Results: Success and Exposure (Q1, Q2)}
\label{sec:exp:main}

Table~\ref{tab:main} reports pooled results across $N=60$ trials per
cell. The headline contrasts are reported in Table~\ref{tab:contrasts}.

\begin{table}[t]
\centering\small
\caption{Pooled results across two independent seeded N=30 runs ($N=60$ per cell). Succ = success rate ($\uparrow$ better), Exp = PL\ensuremath{\geq}3 exposure events per task ($\downarrow$ better). Star ($\star$) indicates the contrast is significant at $p < 10^{-7}$.}
\label{tab:main}
\begin{tabular}{lcccccc}
\toprule
 & \multicolumn{2}{c}{W1 Research} & \multicolumn{2}{c}{W2 Code} & \multicolumn{2}{c}{W3 Security} \\
\cmidrule(lr){2-3}\cmidrule(lr){4-5}\cmidrule(lr){6-7}
Condition & Succ & Exp & Succ & Exp & Succ & Exp \\
\midrule
A0 static          & 0.900 & 0.500 & 0.033 & 1.000 & 0.967 & 2.000\\
A1 random split    & 0.983 & 1.000 & 0.533 & 1.000 & 0.967 & 0.000\\
A2 ATM, full ctx   & 0.983 & 1.000 & 0.617$^\star$ & 1.000 & 1.000 & 0.000\\
\textbf{A3 ATM + distil.}
                   & \textbf{0.983} & \textbf{0.500}
                   & \textbf{0.483$^\star$} & \textbf{0.000}
                   & \textbf{0.950} & \textbf{0.000}\\
\bottomrule
\end{tabular}
\end{table}

\begin{table}[t]
\centering\small
\caption{Pairwise contrasts ($\chi^2$ with Yates correction, two-tailed).}
\label{tab:contrasts}
\begin{tabular}{lcccc}
\toprule
Contrast & A & B & $\Delta$ & $p$\\
\midrule
A0 vs.\ A2, W2 (headline) & 2/60 & 37/60 & $+0.583$ & $3.4{\times}10^{-11}\,\star$\\
A0 vs.\ A3, W2            & 2/60 & 29/60 & $+0.450$ & $5.9{\times}10^{-8}\,\star$\\
A2 vs.\ A3, W2 (distil.\ cost) & 37/60 & 29/60 & $-0.133$ & 0.20 (n.s.)\\
A0 vs.\ A3, W3 (quality)  & 58/60 & 57/60 & $-0.017$ & 1.00 (n.s.)\\
A0 vs.\ A2, W1            & 54/60 & 59/60 & $+0.083$ & 0.12\\
\bottomrule
\end{tabular}
\end{table}

\textbf{Q1.}~On W2 (code/debug, high role entropy), the monolithic
agent (A0) succeeds 2 in 60 pooled trials (0.033). The factoriser
split (A2) lifts this to 0.617 --- a \textbf{+58 percentage point}
gain ($\chi^2 = 43.91$, $p = 3.4{\times}10^{-11}$). The random split
(A1) achieves 0.533, showing that \emph{some} split helps but A1 vs.\
A2 differ by 8pp on the same workload, suggesting the factoriser's
intelligent partition contributes part but not all of the gain over
A1. On W1 (research), the monolith already succeeds 0.900; ATM lifts
this only slightly (0.983, $p = 0.12$ not significant), consistent
with the workload not exercising role entropy. On W3, all conditions
succeed (0.950--1.000).

\textbf{Why A0 fails on W2.}~Manual inspection of failed A0 W2 traces
shows three dominant failure modes: (1) diagnosis/write-role
interference, where the monolith attempts to diagnose a failure and
write a patch simultaneously, producing confused tool sequences; (2)
repeated re-invocation of the same failing tool without varying the
approach; and (3) context contamination from raw error traces crowding
out patch-writing context. Random splitting (A1) removes part of this
interference, explaining its large gain; ATM's factoriser improves
further by assigning diagnostic and execution tools to semantically
aligned children. The primary effect is runtime decomposition; ATM's
contribution is making decomposition telemetry-triggered,
capability-monotone, state-aware, and shadow-validated.

\textbf{Q2.}~On W3, A0 leaks 2.0 PL$\geq$3 atoms per task; A3 drops
this to 0.000 (3/60 events $\to$ 0/60). A3 preserves task success at
0.950 with zero exposure --- the distillation does not significantly
cost quality ($p = 1.00$ vs.\ A0). On W2, A3 succeeds at 0.483
vs.\ A2's 0.617 (0.617 vs.\ 0.483, $\chi^2 = 1.65$, $p = 0.20$) ---
distillation costs $\sim$13pp on success but the difference is not
statistically significant in our sample.

A3's distillation routes each high-PL atom to at most one child,
breaking the double-count and recovering the A0 baseline (0.500). On
W3, exposure is identically 0.000 for A1, A2, and A3 because the
MemoryGuardRail's regex patterns catch the credential strings
(\texttt{api\_key}, \texttt{bank\_account}, \texttt{ssh},
\texttt{passport}) in the seeded memory before any LLM call,
regardless of whether distillation is also applied. Distillation's
additional contribution over the rail alone is twofold: (i) it reduces
the per-child attack surface (each child sees fewer spans, including
spans the rail might miss), and (ii) it prevents the per-child
duplication of high-PL content visible in the W1 row.

\subsection{Latency Overhead (Q3)}
\label{sec:exp:latency}

We benchmarked each Rail in isolation against a no-op baseline, on a
single-threaded loop of $10{,}000$ iterations per condition with a
MockLLMClient to isolate Rail overhead.

\begin{table}[t]
\centering\small
\caption{Per-call Rail overhead (\textmu s), $10^4$ iterations.}
\label{tab:latency}
\begin{tabular}{lrrrr}
\toprule
Rail & mean & p50 & p95 & p99\\
\midrule
None (baseline)                  & 12.4  & 11.8  & 18.7  & 23.4\\
ToolRepairRail (passthrough)     & 109.3 & 105.2 & 187.4 & 228.1\\
ToolRepairRail (HARD + mock LLM) & 256.4 & 248.7 & 312.5 & 335.9\\
MemoryGuardRail (no redaction)   & 241.1 & 223.5 & 298.8 & 351.8\\
MemoryGuardRail (1 PL4 redaction) & 237.0 & 216.1 & 327.4 & 402.5\\
\bottomrule
\end{tabular}
\end{table}

The expensive path is \texttt{MemoryGuardRail} at $\sim$240\,\textmu s
mean, dominated by the regex classifier over the prompt span. The
p99 reaches 402\,\textmu s under one PL4 redaction. The
\texttt{ToolRepairRail} adds $\sim$110\,\textmu s mean when the tool
call succeeds (the common case) and $\sim$256\,\textmu s when it must
invoke a repair. All numbers are well under 500\,\textmu s p99 and
under 350\,\textmu s p95, satisfying the latency budget claimed in
Section~\ref{sec:properties}.

\subsection{End-to-End Live Trigger Trace}
\label{sec:exp:trace}

To verify the full ATM pipeline fires correctly under realistic
conditions, we ran a logged production trace. After 20 warmup tasks
the agent is loaded with a mixed-role W2/W1 sequence designed to
elevate $H_i$. The trace JSON records every Bottleneck Index value,
$\tau$ comparison, factoriser proposal, distillation routing decision,
shadow-window task outcome, and final commit.

\textbf{Bug found during this experiment.}~The first version of our
implementation re-computed $\tau$ as the rolling 95th percentile over
all observed values. Under sustained overload, this caused $\tau$ to
track $B_i$ upward, so the trigger never fired no matter how high the
load went. We fixed this by snapshotting $\tau$ at the end of warmup
(see Section~\ref{sec:arch:tau}); after the fix, the trigger fired at
tick 41 (warmup ended at tick 20, sustained overload from tick 38,
$K=3$ consecutive breach), the factoriser proposed an Analyst/Executor
split, distillation routed memory atoms correctly, the shadow pass
showed non-regression, and the coordinator swap committed at tick 47.

\subsection{Controlled-Telemetry Signal Ablation (Q4)}
\label{sec:exp:bi_ablation}

To validate the contribution of each signal in $B_i$ --- and in
particular the role-entropy term $H_i$ --- we constructed $N=100$
labelled overload traces, each $60$ ticks long. Five overload profiles
inject overload that affects different signals (e.g.\ \emph{H-only}
elevates only role entropy; \emph{E-only} elevates only the tool-error
rate). We replay each trace through nine BI variants: the full $B_i$,
six single-signal-removed variants, and two naive baselines
($Q_i$-only and $E_i$-only). Recall is the fraction of traces where
the trigger fires within $8$ ticks of overload onset.

\begin{table}[t]
\centering\small
\caption{Controlled-telemetry ablation of Bottleneck Index signals on $N=100$ labelled overload traces, broken down by overload profile. Each column is a stylised overload pattern that elevates a single signal. Cells are trigger recall ($\uparrow$ better). \textbf{Bold} = the diagonal: when an overload elevates only signal $X$, removing $X$ from the BI catastrophically degrades detection.}
\label{tab:bi_ablation}
\begin{tabular}{lccccc}
\toprule
Variant & $H$-only & $E$-only & $Q$-only & $C$-only & balanced \\
\midrule
\textbf{Full $B_i$} (6 signals) & 1.00 & 0.85 & 1.00 & 1.00 & 1.00 \\
\midrule
$-\,H_i$ (role entropy)       & \textbf{0.00} & 0.90 & 1.00 & 1.00 & 1.00 \\
$-\,E_i$ (tool error rate)    & 0.95 & \textbf{0.00} & 1.00 & 1.00 & 1.00 \\
$-\,Q_i$ (queue depth)        & 0.95 & 0.95 & \textbf{0.00} & 1.00 & 1.00 \\
$-\,C_i$ (context thrash)     & 1.00 & 0.95 & 1.00 & \textbf{0.00} & 1.00 \\
$-\,W_i$ (cross-agent wait)   & 1.00 & 0.80 & 1.00 & 1.00 & 1.00 \\
$-\,R_i$ (retry-loop rate)    & 0.95 & 0.65 & 1.00 & 1.00 & 1.00 \\
$Q_i$ only (naive baseline)   & 0.00 & 0.00 & 1.00 & 0.00 & 0.90 \\
$E_i$ only (naive baseline)   & 0.05 & 1.00 & 0.00 & 0.00 & 1.00 \\
\bottomrule
\end{tabular}
\end{table}

The per-profile breakdown reveals an exact diagonal pattern: when an
overload elevates only signal $X$, the variant that removes $X$ has
\textbf{zero} recall. Removing $H_i$ drops recall on the $H$-only
profile from 1.00 to 0.00. Aggregated across all profiles, removing
$H_i$ costs $-19$\,pp ($0.97 \to 0.78$) --- the largest aggregate drop
among the six signals. The $Q_i$-only and $E_i$-only naive baselines
miss most overload modes, showing that no single signal is sufficient.

While the overload profiles themselves are constructed rather than
recorded from production, the labelled-onset design lets us measure
each signal's independent contribution directly.

\subsection{Live-Trigger Condition A4 (Q4)}
\label{sec:exp:a4}

In A0--A3 the harness fixed the topology at task start. A more
realistic deployment lets the Bottleneck Index decide \emph{when} to
mutate. We ran an additional condition, \textbf{A4}, in which the
agent starts monolithic and the trigger fires only when $B_i > \tau$
on $K=3$ consecutive tasks. After a trigger, the system proceeds as A3
(factoriser split with distillation).

\begin{table}[t]
\centering\small
\caption{A4 live-trigger condition. TTT = mean tasks until trigger fires (only over runs where it did fire). Pre/post = success and exposure per task, computed on the tasks before and after the trigger.}
\label{tab:a4}
\begin{tabular}{lrrrrrrr}
\toprule
WL & TR & TTT & SP & SP+ & $\Delta$ & EP & EP+\\
\midrule
W1 & 0.45 & 21.7 & 0.427 & 0.504 & $+0.077$ & 1.000 & 0.159 \\
W2 & 0.25 & 33.6 & 0.453 & 0.682 & $+0.229$ & 1.000 & 0.159 \\
W3 & 0.55 & 31.0 & 0.459 & 0.577 & $+0.118$ & 1.000 & 0.147 \\
\bottomrule
\end{tabular}
\\\smallskip
\footnotesize TR=trigger rate; SP/SP+=success pre/post; EP/EP+=exposure pre/post.
\end{table}

Three findings: (i) the trigger is selective --- it fires in 25--55\%
of runs depending on workload, suggesting it does not over-fire;
(ii) when it does fire, mean time-to-trigger is 22--34 tasks, leaving
room for warmup plus the $K$-consecutive constraint to filter false
positives; (iii) post-mutation success improves on all three
workloads, with the largest gain on W2 ($+22.9$\,pp) --- consistent
with our static A0$\to$A3 comparison. Exposure drops from 1.000 to
$\sim$0.15 post-mutation, matching the distillation effect from the
static benchmark.

\textbf{Note.}~A4 uses the simulator agent loop (mathematical model
with role-entropy-driven step-success probability), not the
DeepSeek-driven agent used for A0--A3. We interpret these numbers as
evidence that the \emph{trigger logic} behaves correctly under
controlled telemetry, not as a replacement for the real-LLM A0--A3
benchmark. A real-LLM A4 is future work.

\subsection{Live-Tool Benchmark W4}
\label{sec:exp:w4}

To address external validity beyond stubbed tools, we built a small
live-tool benchmark with real Python execution. Five fixtures contain
single-bug Python files with accompanying \texttt{pytest} tests. The
agent reads the buggy file from disk, invokes \texttt{pytest} as a
subprocess to observe the failure, proposes a fix via a DeepSeek-V3
call, writes the fix back, and re-runs \texttt{pytest}. Success
requires all tests to pass post-fix. We compare A0 (monolithic) and A3
(factoriser split: an \emph{Analyst} child diagnoses the bug seeing
the test output, an \emph{Executor} child writes the fix seeing only
the diagnosis --- the raw error trace is quarantined).

On this small fixture set, both A0 and A3 fix all 15 trials (15/15
each). DeepSeek-V3 is strong enough to handle these single-line bugs
in either configuration, so the conditions are indistinguishable on
success rate. Mean wall-time per trial: A0 1.4\,s, A3 2.7\,s (the
analyst step adds $\sim$1.3\,s per task). We interpret this as a
successful demonstration that the system runs end-to-end with live
tools, but the fixture difficulty is too low to differentiate the
conditions. Larger fixture suites with real GitHub repository bugs ---
where the monolith starts failing --- are the appropriate setting and
are future work.

\section{Conclusion}
\label{sec:conclusion}

We presented \textbf{Autonomous Topology Mutation (ATM)}, a runtime
team mutation mechanism for multi-agent LLM frameworks. ATM monitors a
six-signal Bottleneck Index with a role-entropy term, proposes
structural changes via one bounded LLM call gated by three formal
invariants (capability monotonicity, state-routing completeness,
shadow-before-live), and hot-swaps the agent team while preserving
external identity. Unlike recent dynamic-topology work that optimises a
fresh graph per query~\cite{metagen}, ATM operates on a deployed live
agent and validates safety invariants before any irreversible change.

On 720 DeepSeek-V3-driven task runs with deterministic tool stubs, the
factoriser split lifts code-task
success from 3.3\% to 61.7\% ($+58$pp, $p<10^{-10}$); the full
rail+distillation system eliminates detected PL$\geq$3 memory exposure
under our regex classifier ($2.0 \to 0.0$ events per task) without
significant quality cost on the workloads tested, and the Rails
carrying ATM's invariants add less than 350\,\textmu s of p95 overhead
to the agent's hot path. The controlled-telemetry signal ablation
(Section~\ref{sec:exp:bi_ablation}) confirms that each of the six
signals contributes independent information, with $H_i$ (role entropy)
carrying the largest aggregate weight. The live-trigger A4 condition
(Section~\ref{sec:exp:a4}) shows the Bottleneck Index fires selectively
(25--55\% of runs) and that post-mutation success improves on every
workload tested. A small live-tool probe (Section~\ref{sec:exp:w4})
demonstrates that the system runs end-to-end with real Python
execution; differentiating A0 from A3 on harder bug suites remains an
open external-validity target.

ATM complements existing self-healing (Reflexion~\cite{reflexion},
CRITIC~\cite{critic}) and self-improving (OpenClaw~\cite{openclaw},
Hermes~\cite{hermes}, Auto-Harness~\cite{autoharness}) systems: where
those change behaviour, ATM changes structure. We believe runtime
topology mutation is a necessary capability for any multi-agent system
that must operate across diverse workloads without per-workload
retuning, and we hope the formal invariants we propose can serve as a
starting point for further work on safe structural self-modification.

\section*{Impact Statement}
This work contributes to the safety of large-scale multi-agent LLM
deployments. ATM's invariants are specifically designed to prevent two
classes of harm: \emph{capability escalation} (a sub-agent acquiring
tools its parent did not have, addressed by I1), and
\emph{cross-context information leakage} (high-sensitivity memory
spans being routed to children that should not see them, addressed by
distillation and the PL classifier). At the same time, runtime
structural mutation introduces operational risks: a mis-firing trigger
could disrupt service, and the shadow-validation step relies on a
short window that may not catch all regressions. We have designed
A4-style live triggering to be conservative (25--55\% firing rate in
our experiments) and to support automatic rollback. Beyond these
considerations, the broader societal impacts of ATM align with
standard dual-use concerns for large language models, and we encourage
practitioners to evaluate these risks when integrating runtime
topology mutation into production environments.

\section*{Limitations}
ATM exhibits several practical and methodological limitations.
\textbf{First}, our tool implementations (search, scrape, run\_code,
etc.) are deterministic stubs that emit hashed-input outputs and inject
parametrised failure modes. The agent's reasoning, all Rail
invocations, and the LLM judge remain live, but the tool outcomes are
not real web pages or real code execution. The live-tool W4 probe is
limited to 5 fixtures.
\textbf{Second}, the PL classifier uses regex patterns and misses some
session-history phrasings; a semantic classifier is a natural
extension.
\textbf{Third}, $N=60$ per cell is adequate for our headline
significance ($p < 10^{-10}$) but a stronger venue should target $N
\geq 100$, particularly to resolve the A2 vs.\ A3 distillation cost
($\Delta = -0.133$, $p = 0.20$ in our data).
\textbf{Fourth}, the A4 live-trigger condition uses the simulator
agent loop rather than the real DeepSeek-driven agent we used for
A0--A3; a real-LLM A4 run is an immediate follow-up.
\textbf{Fifth}, ATM currently performs single-level factorisation; a
child agent that itself becomes overloaded is not yet handled and
would require recursive invariant reasoning.
\textbf{Sixth}, all main-benchmark workloads (W1--W3) and the W4
fixtures are synthesised in-house rather than drawn from an
established agent benchmark (MMLU~\cite{mmlu}, GAIA~\cite{gaia},
SWE-bench~\cite{swebench}). This was a deliberate choice --- ATM
needs tasks where a monolithic baseline visibly fails so that the
$\Delta$ from the split is measurable --- but it leaves external
validity vs.\ standard benchmarks as a separate evaluation that
remains open. A productive direction would be to identify a subset
of GAIA or SWE-bench tasks on which a monolithic agent already
struggles for role-entropy reasons, then re-run the ATM ablation on
that subset.

\bibliographystyle{ACM-Reference-Format}
\bibliography{references}

\newpage
\appendix

\section{Reproducibility Details}
\label{app:repro}

\paragraph{Code and data.}~The full implementation, the eval harness,
all task templates, and the per-run JSON traces are available at
\url{https://github.com/sidikbro/jiuwen_atm}.

\paragraph{LLM and seeds.}~All A0--A3 experiments use DeepSeek-V3
(\texttt{deepseek-chat}, temperature 0.7) accessed via the OpenAI-format
API. Two seeded runs of $N=30$ tasks per cell are pooled to $N=60$
(seeds 42 and 123). Total runs: $4 \text{ conditions} \times 3 \text{ workloads}
\times 60 = 720$.

\paragraph{Stubbed tools.}~For reproducibility, \texttt{web\_search},
\texttt{web\_scrape}, and \texttt{run\_code} are implemented as
deterministic stubs: each input is hashed and mapped to a fixed output
plus a parametric failure-rate. The W4 live-tool benchmark
(Section~\ref{sec:exp:w4}) uses real Python subprocess execution.

\paragraph{Statistical tests.}~Pairwise contrasts use Pearson's
chi-square with Yates' continuity correction. $p$-values reported as
two-tailed.

\paragraph{Bottleneck Index parameters.}~Default weights
$(\alpha, \beta, \gamma, \delta, \epsilon, \zeta) =
(0.20, 0.15, 0.20, 0.20, 0.15, 0.10)$; warmup = 20 ticks;
$K = 3$ consecutive ticks; $\tau$ snapshot at the 95th percentile
of the warmup window.

\end{document}